\documentclass[times]{TRR}

\usepackage{moreverb,url}
\usepackage{mathtools}
\usepackage{graphicx}

\usepackage[colorlinks,bookmarksopen,bookmarksnumbered,citecolor=red,urlcolor=red]{hyperref}

\newcommand\BibTeX{{\rmfamily B\kern-.05em \textsc{i\kern-.025em b}\kern-.08em
T\kern-.1667em\lower.7ex\hbox{E}\kern-.125emX}}

\def\volumeyear{2020}

\begin{document}


\runninghead{Zare et al.}

\title{Modular Self-Lock Origami:  design, modeling, and simulation to improve the performance of a rotational joint}

\author{Samira Zare\affilnum{1},
    Alex Spaeth\affilnum{1},
    Sandya Suresh\affilnum{2},
    and Mircea Teodorescu\affilnum{1}}

\affiliation{
    \affilnum{1}Department of Electrical and Computer Engineering, University of California Santa Cruz\\
    \affilnum{2}SIP program, University of California Santa Cruz}

\corrauth{Samira Zare \\
    Department of Computer Science and Engineering \\
University of California Santa Cruz}
\email{szare@ucsc.edu}

\begin{abstract}
Origami structures have been widely explored in robotics due to their many potential advantages. Origami robots can be very compact, as well as cheap and efficient to produce. In particular, they can be constructed in a flat format using modern manufacturing techniques. Rotational motion is essential for robotics, and a variety of origami rotational joints have been proposed in the literature. However, few of these are even approximately flat-foldable. One potential enabler of flat origami rotational joints is the inclusion of lightweight pneumatic pouches which actuate the origami’s folds; however, pouch actuators only enable a relatively small amount of rotational displacement. The previously proposed Four-Vertex Origami is a flat-foldable structure which provides an angular multiplier for a pouch actuator, but suffers from a degenerate state. This paper presents a novel rigid origami, the Self-Lock Origami, which eliminates this degeneracy by slightly relaxing the assumption of flat-foldability. This joint is analysed in terms of a trade-off between the angular multiplier and the mechanical advantage. Furthermore, the Self-Lock Origami is a modular joint which can be connected to similar or different joints to produce complex movements for various applications; three different manipulator designs are introduced as a proof of concept.
\end{abstract}

\keywords{
Origami, Rotational joint, Deployable structure, Modular, Crease pattern, Earwig wing, Miura-ori, Spherical mechanisms, Pouch motors, Manipulator
}

\maketitle


\section{Introduction}


Origami structures have the potential to enhance robotics in various ways. They can help save space \cite{jasim2018origami,tang2014origami,arya2017crease}, reduce energy consumption \cite{zhai2020situ,quaglia2014balancing,ye2022novel}, decrease production time and cost \cite{dai2010origami,onal2011towards,zhakypov2015design,yang2017smartphone,kimionis20153d} by offering a compact and lightweight structure, and having a simple flat structure that is easy to produce. Hence, by replacing conventional robotic structures with origami structures, robots could take advantage of these benefits. This paper focuses on an origami rotational joint structure that could substitute the traditional revolute joint \cite{norton2008design} while offering the advantages of origami.

A variety of origami robots, ranging from legged walkers \cite{rus2018design,mehta2014cogeneration,mehta2014end,kohut2013precise,haldane2013animal,zhakypov2018design} to grippers and more \cite{rossiter2014kirigami, firouzeh2017grasp, geckeler2022bistable, chan2017design, suzuki2020origami}, make use of the simple fold as a rotational element in order to build up more sophisticated robots.
The simple fold is ideal for these purposes because it is simple and compact.
Tendon-driven robogami \cite{firouzeh2017under} joints with adjustable stiffness using SMP (shape memory polymer) layer could rotate along multiple axes. The joints are flat-foldable and modular. However, the rotational motion has limited range which depends on the solid panels' thickness and the flexible joint between the panels.

Many rotational joints have been developed beyond the simple fold; some can create limited rotation in multiple different directions \cite{koh2012omega,boyvat2017addressable,salerno2016novel}.
Pneumatic origami rotational actuators can produce significant force, but are large and cannot be flat-folded for deployability \cite{yi2018customizable}.
Foldable designs for hinge and pivot joints have been proposed \cite{sung2015foldable}.
These joints can be combined to generate desired kinematics, but they are relatively complicated and cannot be flat-folded.
Curved patterns can add stiffness to an origami structure capable of rotational motion \cite{zhai2020situ,taylor2019mr,baek2020ladybird,saito2017investigation}.
A curved kirigami joint has been shown to achieve rotational motion up to almost $17^\circ$ in multiple directions \cite{qiu2021design}. In general, curved surfaces generate a low range of rotational motion compared to simple folds. Also, the fold action is impossible due to their geometrical constraints (no fold line).

\begin{figure}
\centering
\includegraphics[width=\linewidth]{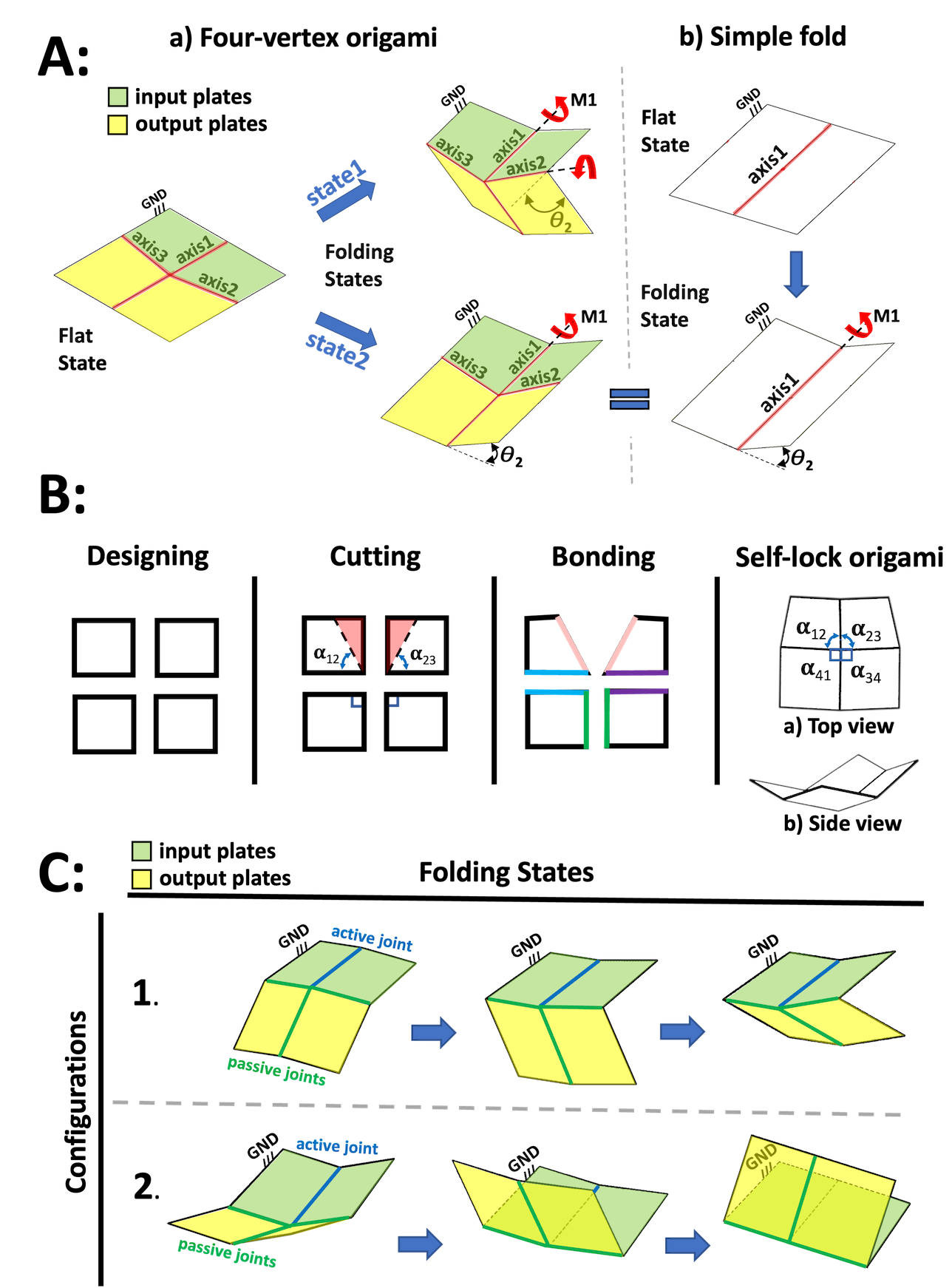}
\caption{
{\bf A}:
a) Transformation from the flat state of the Four-Vertex Origami to the two possible folded states for the same input. The origami in state 1 undergoes rotation at all four joints, while in the degenerate state 2, axes 1 and 4 remain collinear. The input plates, covered by the pouch actuator, are coloured green, while the output plates, with no actuator involvement, are coloured yellow.
b) Transformation from the flat state of the Simple Fold to the folded state, equivalent to state 2 of the Four-Vertex Origami.
{\bf B}: Design and Modelling of the Self-Lock origami. Design: drawing the square plates. Cutting: reducing the central angles on two of the plates. Bonding: defining revolute joints between corresponding edges of adjacent plates. Self-Lock origami: the final product in different views, with two reduced central angles $\alpha_{12} = \alpha_{23}$, and two central angles $\alpha_{34} = \alpha_{41} = 90^\circ$.
{\bf C}: Transition from almost flat state to a folded state in configurations 1 (down) and 2 (up). The blue active joint is actuated by the pouch actuator, while the three green passive joints move based on the rotation of the active joint.
}\label{probFabricMov}
\end{figure}

A potential candidate to substitute the traditional rotational linkage in an origami robot is the Four-Vertex Origami, which generates 8.4 times more rotational angle than a simple fold for a given actuator displacement \cite{zare2021design}.
This is crucial when using pouch actuators, which pair naturally with flat-foldable or semi-flat-foldable origami due to their flat shape when deflated, but have a limited range of angular movement \cite{niiyama2015pouch}.
It can be manufactured in different shapes and sizes all the way down to the thickness of a sheet of paper, allowing its weight and size to be adjusted based on the application.
Furthermore, it can be assembled with other origami joints to create complex movements \cite{brown2022approaches}.
However, such an origami is underactuated when folded from flat, and in fact has a degenerate state equivalent to a simple fold (figure \ref{probFabricMov}A).
The uncertainty of which fold line will be activated when the pouch motor is inflated increases the chance of failure in the joint's rotation.

Faber et al.\cite{faber2018bioinspired} designed an earwig-inspired spring origami joint similar to the four-vertex origami but with central angles around the common vertex summing to less than $360^\circ$.
The flexibility and extensibility of this joint provides a self-lock mechanism during the insect's flight.
However, there is an inherent trade-off in the use of non-rigid origami:
the joint spring must be matched precisely to the task,
and the stretchability of the fold lines declines over time and with every use.
Furthermore, depending on the number of folds, the width of flexible fold lines could be large, resulting in a weaker structure.

This paper proposes a new rigid origami design, the Self-Lock Joint, that combines the Four-Vertex origami \cite{zare2021design} and the earwig-inspired joint \cite{faber2018bioinspired}.
Reducing the central angles of the origami eliminates the degenerate state of the Four-Vertex origami.
Combining with the assumption of a fold line with non-extensible material and zero gaps, the joint could have a deployable structure, save energy, provide a state-lock mechanism, great moment, and downward and upward rotational movements of the structure.


\section{Design}
\label{design}

The Self-Lock Origami is a rigid origami whose central angles sum to less than $360^\circ$.
The design consists of four solid adaptable plates connected by four joints (figure \ref{probFabricMov}C), three of which are passive and one of which will be actuated.
Figure \ref{probFabricMov}B shows the design and modelling process.
First, square-shaped origami plates of side length 25mm are defined. For modelling purposes, these are treated as having zero thickness.
Then, the central angles of two of the plates are reduced by drawing a line from one corner of the square (figure \ref{probFabricMov}B) with a specified angle $\alpha$ and extending the line until it intersects with the other edge of the square.
Revolute joints were then defined at the edges of adjacent plates, indicated with edge colour in
 figure \ref{probFabricMov}B.
These revolute joints are a first-order approximation to the kinematics of a flexible fold line.

\subsection{Motion Simulation}\label{simulation}

For origami assembly, joint constraints are necessary.
Various shapes are created by cutting angles in different origami plates' positions:
any of the origami's four central angles can be reduced in order to achieve the self-locking property.
Although figure \ref{probFabricMov} demonstrates only one of these, 16 different configurations were considered: four where the central angle was subtracted from both sides of a single fold line (figure \ref{L11}), and twelve other configurations shown in the supplemental material.
Note that only cuts reducing the central angles are considered, as an origami could have various shapes around its edges without compromising its motion \cite{zare2021design}.

\begin{figure}
\centering
\includegraphics[width=\linewidth]{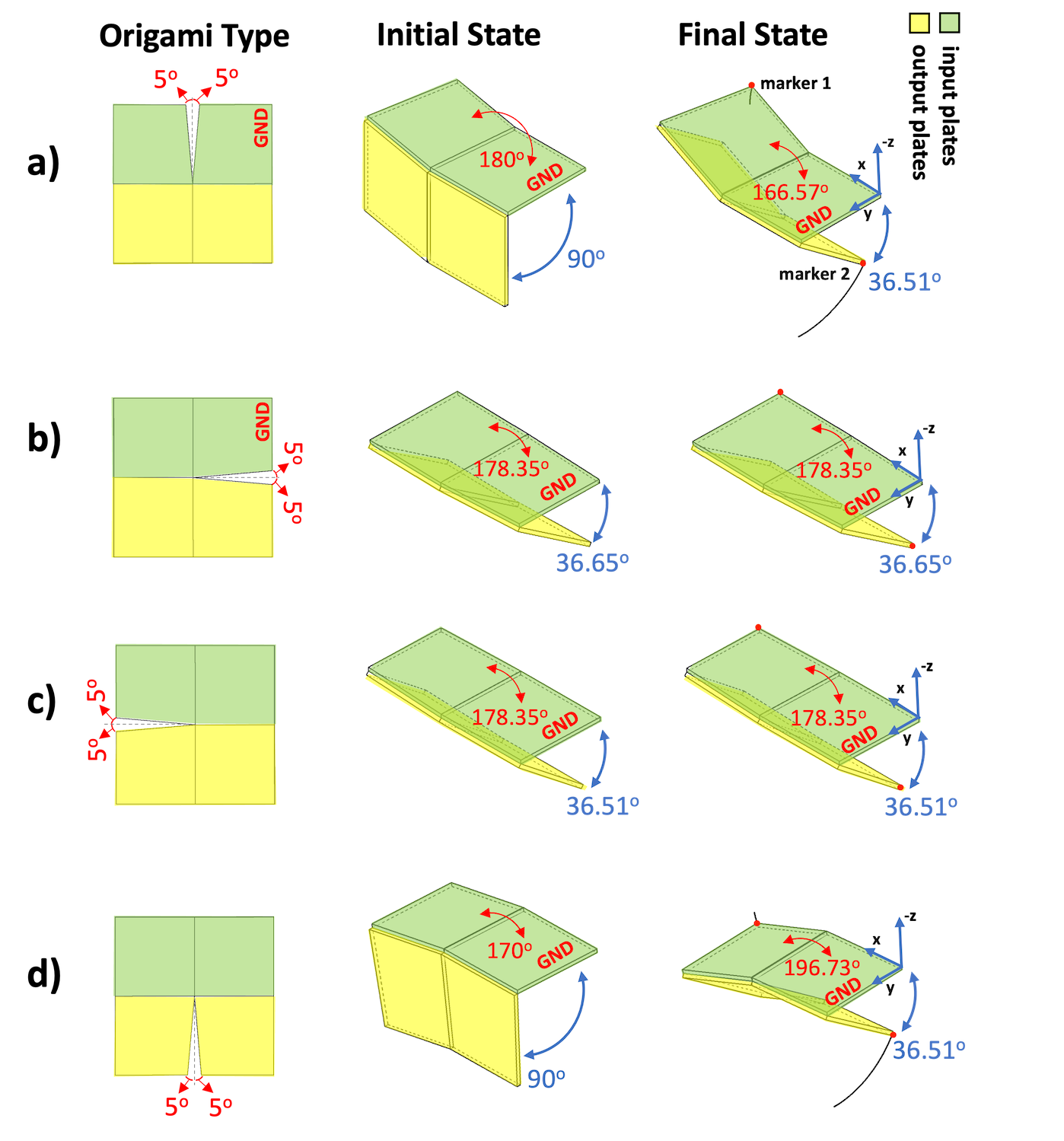}
\caption{
Four different possibilities for the reduction of the central angle in the Self-Lock Origami. The input and output plates are highlighted with green and yellow colors, respectively.
Initial state: Closest possible configuration to the flat state of the input plates.
Only origami type a achieves a flat state for the input plate, $\theta_1 = 0^\circ$.
Final state: manually defined Maximum Possible Folding configuration (MPF) between the grounded and the output plate.
} \label{L11}
\end{figure}

The structure of the grounded plate 1, connected to plate 2, is equivalent to a simple fold (figure \ref{motion}a). Therefore, the origami's input angle, $\theta_1$, and input moment are the same as the rotational angle and the output moment of a simple fold structure with the same actuator. Different actuators could be used to increase the input angle limits. However, it is not necessary since the output angle changes most for $\theta_1$ near zero (figure \ref{motion}).
Both the semi-flat and the MPF state occur at values of $\theta_1$ near $90^\circ$.
Thus, in order to obtain the full range of motion, there must be pouch motors on both sides of plates 1 and 2.

Various motion simulations were implemented using the contact solver of Autodesk Inventor 2019 to find an origami that could maintain a flat configuration between its input plates at its initial state and obtain the maximum rotational movement (optimal performance).
Achieving a flat state by the origami's input plates could offer a more simplified model and better control by the pouch motor's actuator.
The pouch motor will be able to maintain a zero pressure state (no inflation) and provide a known shape and dimensions at the initial state for the modelling.

In all simulations, plate 1 is a grounded link at a fixed position with its outer corner at the origin. Gravity is neglected for simplicity and consistency with the analytical model.
The kinematics of the mechanism are simulated by sweeping the input angle between what the ``semi-flat'' and ``maximum possible fold'' (MPF) state.
The semi-flat state is defined as the state where the absolute sum of all the angles between origami plates is at its maximum.
Since the origami models to have zero thickness, they are kinematically capable of achieving fold angles more extreme than a real origami, so the maximum possible fold state is chosen to set the angle $\theta_4$ between the input and output plate to the constant MPF angle $\gamma \coloneqq 36.5^\circ$.

Figure \ref{L11} shows origami types defined by different cutting positions. The green and yellow plates are input and output plates respectively. The ``initial state'' pictured in figure \ref{L11} shows the closest angle that origami's input plates could get to $180^\circ$ (flat form), and the ``final state'' is the MPF state.
The motion of the outer corners of the input and output plate is shown using markers 1 and 2.

In order to obtain an origami which is symmetrical and also capable of reaching a state with fully flat input plates, origami type a is used in the remainder of the paper.
If the flat state of the input plates is not considered, all the origami types provide almost the same amount of rotational motions in slightly different directions.
This is discussed in the supplemental material (together with an expanded version of figure \ref{L11}) using an algorithm described previously \cite{zare2021design}.
Additional fold types not pictured were also considered. Certain of these types could also reach a state with flat input plates, but type a is used due to its symmetry, which simplifies modelling.


\section{Modelling}\label{modelings}

The Self-Lock Origami is a single-vertex origami with four fold lines, meaning that it has a single degree of freedom \cite{hull2002modelling, hull2002combinatorics, song2016microscale}.
As shown in figure \ref{probFabricMov}C, the same components can be assembled into two distinct configurations whose motions are vertical mirrors of each other.
These configurations can be used in applications to achieve motion in two different directions.

In this section, the relations between the origami's angles, output moment, input moment, and the pressure in a pouch actuator are developed.
Pouch actuators are a practical choice for the Self-Lock Origami because they are planar and compact.
They can be treated as a layer of the origami during layer-by-layer construction of an origami robot by programmable machines \cite{niiyama2015pouch}.
The mechanical work of the pouch motor inflation converts into the deformation of its shape and changes in its curvature length. These cause angular motions in the simple fold (a hinged structure) that it has been attached to \cite{niiyama2015pouch}.

\subsection{Origami Structure as a Spherical Mechanism}

The kinematics and dynamics of single-vertex rigid origami can be studied by noting that they are mechanically identical to spherical mechanisms \cite{bowen2014position, zare2021design}, for which a detailed theory exists \cite{chiang1988kinematics}.
To do this, the origami folds between the plates are treated as revolute joints.
Then, since each plate of a single-vertex origami is assumed to be inflexible, the single vertex acts as a fixed centre about which the four plates move as the bars of a spherical four-bar mechanism.

The first quantity of interest is the kinematic relationship between the input angle $\theta_1$ and output angle $\theta_4$.
The kinematics of a spherical four-bar mechanism are determined entirely by the angular lengths $\alpha_{12}$, $\alpha_{23}$, $\alpha_{34}$, and $\alpha_{41}$ of its links.
Given these quantities, the input and output angles are related by the following equation \cite{chiang1988kinematics}:

\begin{multline}
    \cos\alpha_{23} \cos\alpha_{41} \cos\alpha_{12}
    - (\sin\alpha_{23} \cos\alpha_{41} \cos\theta_4
    \\
    + \cos\alpha_{23} \sin\alpha_{41} \cos\theta_1) \sin\alpha_{12}
    \\
    + \sin\alpha_{23} \sin\alpha_{41} (\sin\theta_1\sin\theta_4
    - \cos\theta_1\cos\theta_4\cos\alpha_{12})
    \\
    = \cos\alpha_{34}
\end{multline}

However, the structure of the Self-Lock Origami allows this equation to be simplified significantly.
First, the angles $\alpha_{41}$ and $\alpha_{34}$ are both equal to $90^\circ$, which zeros several terms and simplifies others.
The symmetry of the system allows the parameter $\alpha \coloneqq \alpha_{12} (= \alpha_{23})$ to be defined, so several more terms can be simplified and combined, resulting in a simple relation between $\theta_1$ and $\theta_4$.

\begin{equation}\label{theta4}
\tan\theta_4 = \cos\alpha \cot\left(\frac{\theta_1}{2}\right)
\end{equation}

Besides the relation between the input and output angles, the kinematics of the other angles $\theta_2$ and $\theta_3$ may also be of interest.
Due to the symmetry of the origami, $\theta_2 = \theta_4$, but $\theta_3 \not= \theta_1$.
Instead, $\theta_3$ is calculated using the following relation \cite{yang1964application}:

\begin{equation}\label{theta3}
\cos \theta_{3} = \sin^2 \alpha \cos\theta_1 - \cos^2\alpha
\end{equation}

Up to angular equivalence, the kinematic equations \ref{theta4},\ref{theta3} each have two distinct solution curves, corresponding to the two configurations of the Self-Lock Origami.
In the up configuration, $\theta_4$ ranges over positive angles from near zero to almost $180^\circ$, whereas in the down configuration $\theta_4$ ranges over positive angles from near zero to almost $-180^\circ$.
In computations, the atan2 function is used to ensure that the returned value of $\theta_4$ corresponds to the correct configuration, as the single-argument $\tan^{-1}$ would result in a discontinuous curve.
Using $\cos^{-1}$ to calculate $\theta_3$ returns a single continuous positive curve, so the result is simply multiplied by $-1$ in the down configuration.

\subsection{Input and Output Moments}

\begin{figure}
\centering
\includegraphics[width=\linewidth]{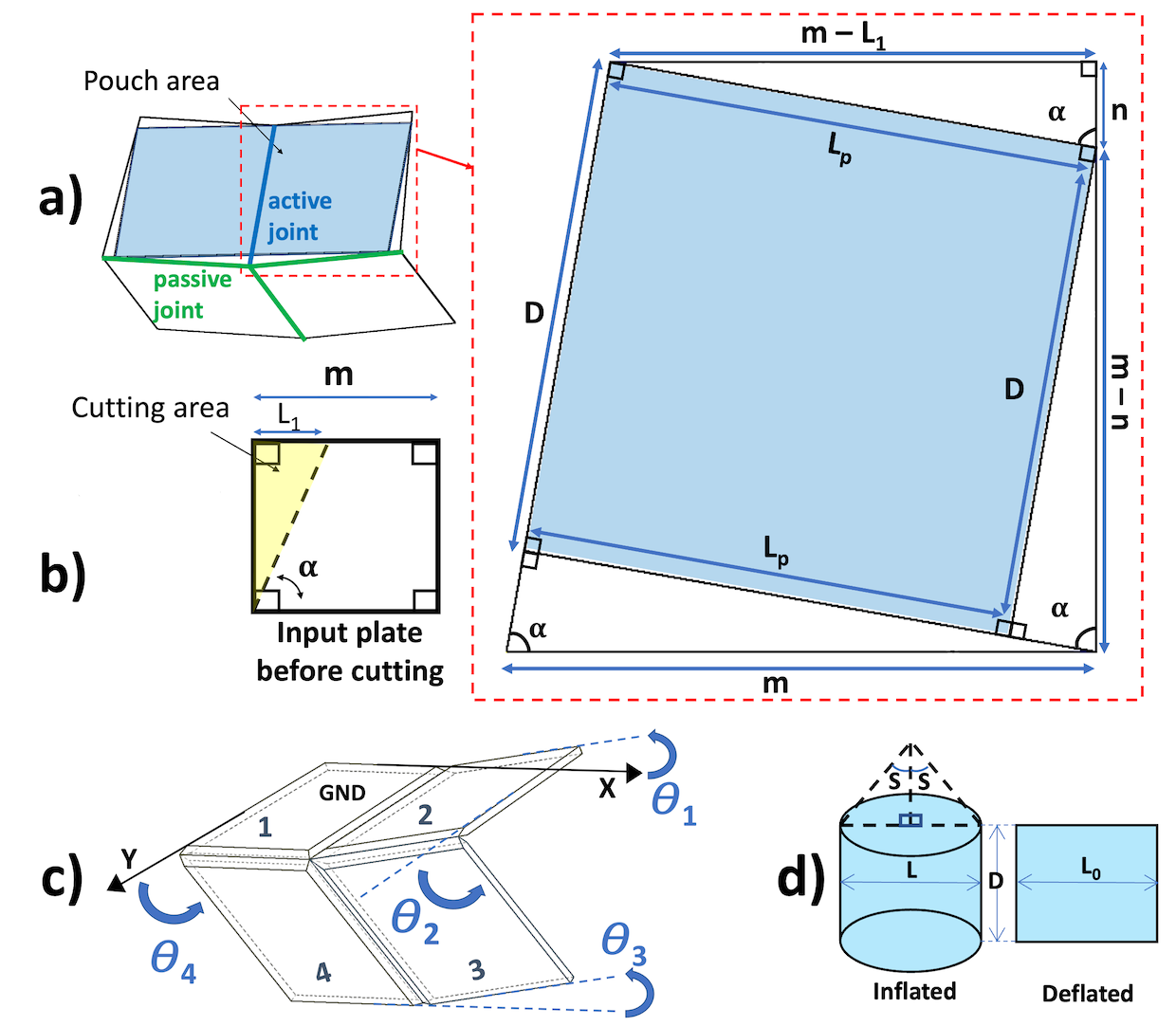}
\caption{
a) Schematic of the self-lock origami with attached pouch motor. The pouch has a rectangular shape before inflation, whereas the plates are trapezoidal. The pouch dimensions $L_p$ and $D$ are derived from the measurements of the plates. The active joint, marked in blue, is a revolute joint defined between the origami plates covered by the pouch actuator. The passive joints, marked in green, do not have contact with the actuator.
b) Input plate with schematic of central angle reduction.
c) Kinematic angles and plate numbers used in analytical modelling of the Self-Lock Origami.
d) Inflated and deflated pouch actuator and its parameters.
} \label{pouch}
\end{figure}

We model the pouch motor as non-extensible with zero bending stiffness.
If fully inflated while detached from the origami, it would take a cylindrical form with height $D$, but it is affixed to the origami in its flat state as shown in figure \ref{pouch}, resulting in a rectangular shape with width $L_0$ and height $D$.
This length is split between plates 1 and 2, so we also define $L_p \coloneqq L_0 / 2$.
The dimensions are chosen to fit the available space on the cut input plates, as shown in figure \ref{pouch}.

The moments that have been applied to (input moment) or generated by (output moment) the origami are modelled under the assumption of constant air pressures on the system's pouch motor. The moments could be calculated using the origami's input angle. A range of $-90^\circ$ to $90^\circ$ degrees has been selected for the input angle, $\theta_1$, due to the physical limitation of the pouch motor to produce even near $90^\circ$ degrees rotational angle. Both upward and downward movements can be created by adding two pouch motors on the origami's input plates:
one on top for $\theta_1:(0^\circ,90^\circ)$ range of motion, and another one on the bottom for $\theta_1:(-90^\circ, 0^\circ)$ range of motion.

To estimate the input moment applied to a simple fold (a hinged structure), Niiyama et al. developed 
equation \ref{T1} \cite{niiyama2015pouch}. The Self-lock origami's input plates have a structure similar to a simple fold, so this equation can be used to calculate its input moment. Based on the proposed origami structure, the equation relies on the origami's input angle $\theta_1$ and $\alpha_{12}$. Given the fixed pressure $P$ as well as the width $D$, half-length $L_p$, and central angle $S$ of the pouch, the input moment can be calculated as follows:

\begin{multline}\label{T1}
        M_\text{input} = \frac{L_p^2 D P}{2S^2}\bigg(-1 + S^2 + \cos 2S \\
                       -\sqrt{2} \cos S \sqrt{-1 + 2S^2 + \cos 2S}\bigg)
\end{multline}

Here $S$ is derived from the assumption that the surface of the pouch acts as a section of a cylinder with constant curvature. The central angle is the angle subtended by the arc of this surface, which depends on $\theta_1$ and can be approximated well for $\theta_1 \in [-\frac\pi2, \frac\pi2]$ as follows \cite{sun2015self}:

\begin{equation}\label{phi}
    \begin{aligned}
        &S(\theta_1) = \sqrt{6(1 - L(\theta_1)/L_0)} \\
        &\text{where }L(\theta_1) = L_0 \sqrt{2(1 + \cos\theta_1)}
    \end{aligned}
\end{equation}

$M_\text{input}$ and $S$ also depend implicitly on the dimensions of the rectangular uninflated pouch.
All four plates of the origami were originally squares of side length $m = 25mm$, but the reduction of central angles to $\alpha$ on the input plates reduces the space available for the pouch, as illustrated in figure \ref{pouch}.
The length of the top side is reduced by a length $L_1 = m \cot\alpha$, and the misalignment of the right angles of the plates and the pouch means that the top right corner of the pouch touches the side of the plate at one point a distance $n = (m - L_1)\cot\alpha$ from its corner. The pouch length $L_p = L_0 / 2$ and width $D$ are then computed geometrically as follows:

\begin{equation}
    \begin{aligned}
        L_p &= (m - L_1) \csc\alpha = m(1 - \cot\alpha) \csc\alpha \\
        D &= (m - n) \csc\alpha \\
          &= m \csc\alpha (1 - \cot\alpha(1 - \cot\alpha))
    \end{aligned}
\end{equation}

Finally, the mechanical advantage of a generic spherical four-bar mechanism has been shown to be calculated as follows (adapted to a different naming convention for central angles) \cite{yang1965static}:

\begin{equation}\label{MechanicalAdvantage}
    \text{M\!A} = \frac{M_\text{output}}{M_\text{input}}
    = \frac{\sin\alpha_{41} \sin\theta_3}{\sin\alpha_{23} \sin\theta_2}
    = \frac{\sin\theta_3}{\sin\alpha \sin\theta_2}
\end{equation}


\section{Model Results}\label{results}

\begin{figure}
\centering
\includegraphics[width=\linewidth]{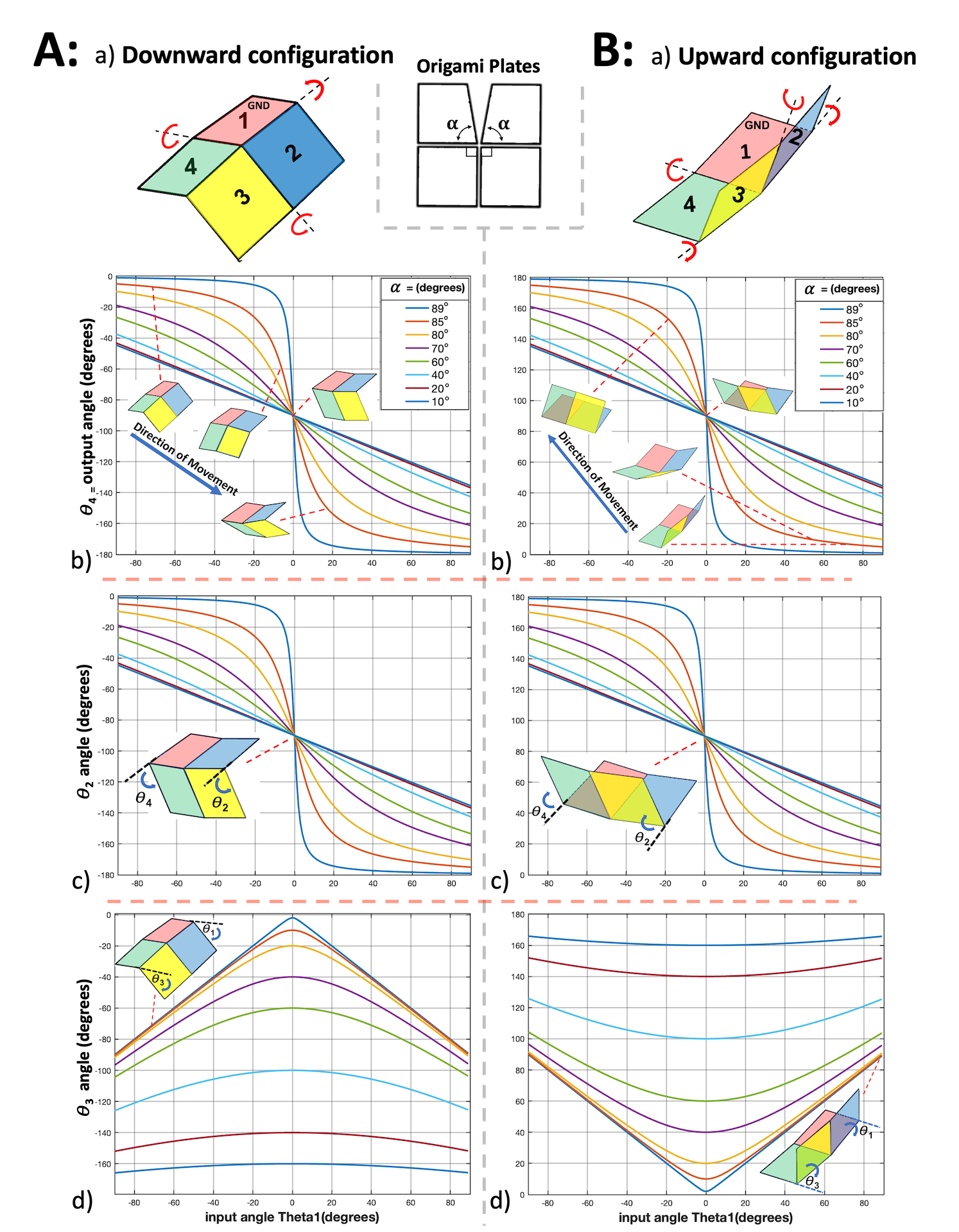}
\caption{
Analytical kinematics of the Self-Lock Origami. Configurations 1 and 2 are shown in columns A and B.
a) Diagram of the origami joint, with plates numbered and colour-coded. As plate 2 moves towards plate 1, plates 3 and 4 experience either downward or upward motion depending on the configuration.
b-d) the kinematic angles $\theta_4$, $\theta_2$, and $\theta_3$ as functions of the input angle $\theta_1$, together with some example states selected from along each curve.
} \label{motion}
\end{figure}

This section demonstrates the kinematics of the Self-Lock Origami as well as its input and output moment, including how these quantities and their relationships change with $\alpha$.
Figure \ref{motion} shows the kinematics of the two configurations of the origami.
Figure \ref{motion}b displays $\theta_4$ as calculated by solving equation \ref{theta4}, figure \ref{motion}c shows $\theta_2$, which is equal to $\theta_4$ by symmetry, and figure \ref{motion}d shows $\theta_3$ by solving equation \ref{theta3}.

The closer $\alpha$ is to $90^\circ$, the steeper the slope of the output angle $\theta_4$ in the regime where $\theta_1$ is near zero.
The total observed range in the output angle $\theta_4$ is about $180^\circ$ regardless of the value of $\alpha$.
The range of $\theta_3$, on the other hand, changes significantly as $\alpha$ decreases.
This is shown in figure \ref{motion}d, where the curves for both configurations for various values of $\alpha$ are symmetrical about local extrema at $\theta_1 = 0$.
As the angular deficit increases, this extreme value moves further from zero, whereas the overall range and the slope of $\theta_3$ decrease.

\begin{figure}
\centering
\includegraphics[width=\linewidth]{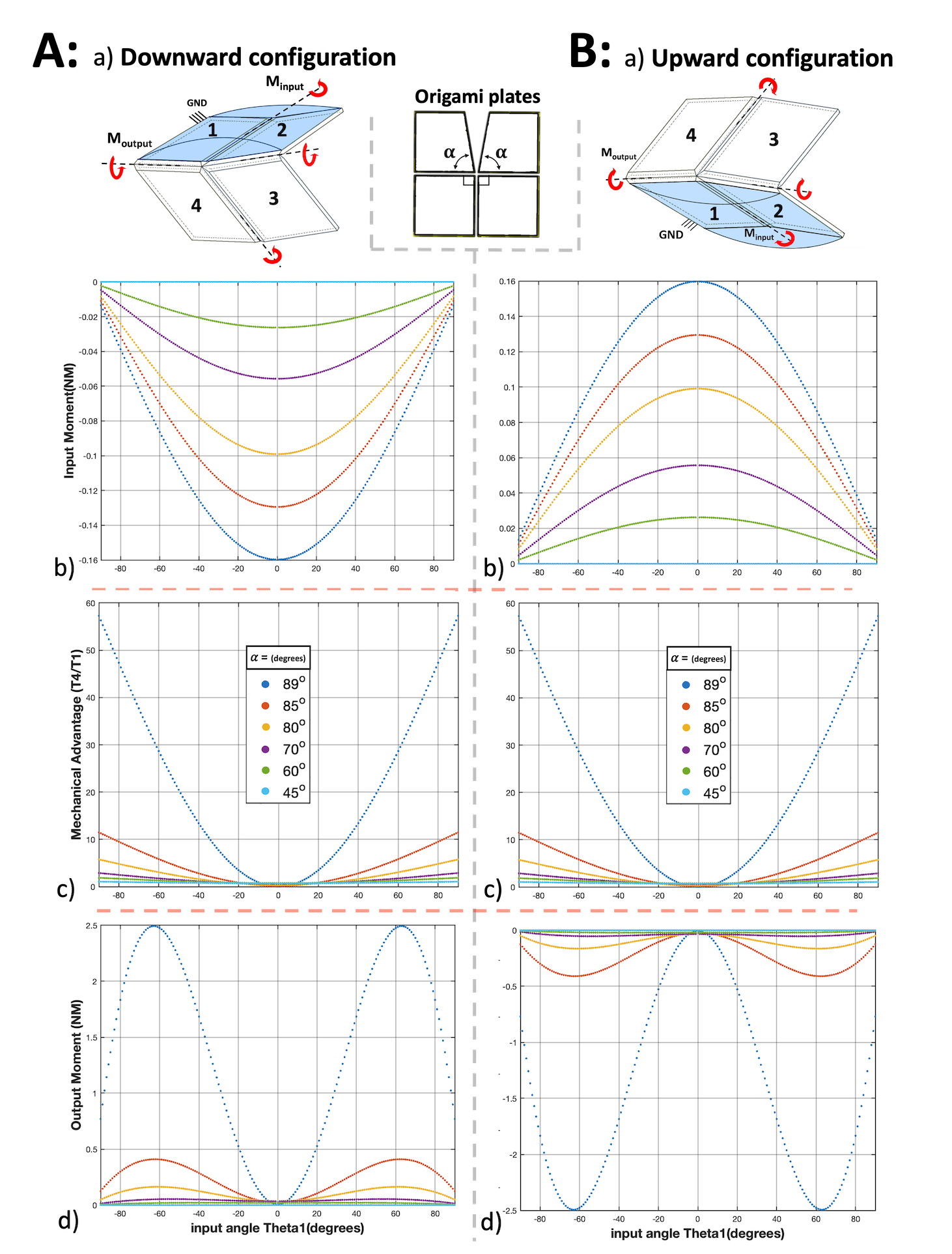}
\caption{
Mechanical advantage of the Self-Lock Origami for a variety of values of $\alpha$. Configurations 1 and 2 are shown in columns {\bf A} and {\bf B}.
a) Diagram of the origami joint with a pouch motor installed on plates 1 and 2. At constant pressure $P = 10kPa$, this produces an input moment $M_{input}$ acting on the input plates. The origami structure converts this to an output moment $M_{output}$.
b) The input moment generated by the pouch motor as a function of the input angle $\theta_1$ according to equation \ref{T1}.
c) Mechanical advantage $ MA = M_{output}/M_{input}$ as a function of $\theta_1$. Although $M_{input}$ is maximised for $\theta_1 = 0$, $MA$ goes to zero.
d) Output moment $M_{output}$ as a function of $\theta_1$.
} \label{moment}
\end{figure}

Figure \ref{moment} illustrates the theoretical curves \ref{T1},\ref{MechanicalAdvantage} of input moment and mechanical advantage as a function of $\theta_1$ for different values of the reduced central angle $\alpha$.
Also depicted is the output moment $M_{output} = MA,M_{input}$.
For the moment study, only one of the origami pouches is considered, since in the typical use case only one pouch will be inflated at a time.
The top origami pouch is activated for the downward movement, and for the upward movement, the bottom pouch is.
Also, the direction of input and output moments are opposite (figure \ref{moment}ABa, Downward and upward movements).

The figure displays curves corresponding to a variety of different values of $\alpha$.
Due to the symmetry between the downward and upward configurations of the Self-Lock origami, both the input and output moment differ only by sign between the two configurations.
As the input angle changes from $-90^\circ$ to $0^\circ$ degrees, the absolute value of the input moment changes from zero to its maximum value at $\theta_1 = 0^\circ$ degrees since the pouch motors work has not been converted into any motion in this state.
As the origami rotates and the input angle gets farther from $0^\circ$ degrees, the absolute value of the input moment decreases.
The shape of this decrease is symmetric about the $\theta_1 = 0^\circ$ line.
These changes are due to observing the maximum rotational motion near the $0^\circ$ degrees input angle and the small input moment values near $\theta_1 = 90^\circ$ or $-90^\circ$.

Note that the figure does not consider the effect of varying the pouch pressure $P$.
The input moment given by equation \ref{T1} depends only linearly on $P$, so all analysis is performed at a constant value $P = 10kPa$.
The higher the pressure, the bigger the input and output moments are.
However, large pressure values could jeopardise the inextensibility assumption for the pouch motor's material and result in model failure.
Thus the limits of this model's applicability are determined by the material properties of the pouch motor, which must be constructed from a material with a high Young's modulus while preserving flexibility properties.

Both downward and upward movements have the same mechanical advantage $MA$ since they are only different in their moment's directions.
As the input angle increases from negative to positive the mechanical advantage decreases and then increases in a concave shape.
The closer plates 1 and 2 get to the flat state ($\theta_1 = 0^\circ$ degrees), the closer the mechanical advantage is to zero since the output moment goes to zero.
Crucially, the mechanical advantage is smallest near $\theta_1 = 0^\circ$, exactly where the slope of $\theta_4$ with respect to $\theta_1$ is greatest.
Therefore, there is a trade-off between the speed and the moment of the origami's movement.
The closer $\alpha$ is to $90^\circ$, the more this trade-off is taken, with the speed of the origami near zero and the output moment far from zero both increasing dramatically.
However, the limiting value is not physically reasonable: at $\alpha = 90^\circ$ the origami does not function.
Similarly, the slope and the mechanical advantage both approach infinity in this limit due to division by zero.
Therefore, in the remainder of the paper, the value of $\alpha$ where not otherwise stated shall be taken to be $89^\circ$, maximising these quantities as well as the flat-foldability of the structure while remaining within a practically realisable regime.


\section{Manipulators}\label{manipulators}

As a proof of concept, three different origami manipulators constructed from combinations of multiple Self-Lock Origami units are presented.
These manipulators can be divided into three categories: Rotational, Translation, and Modular manipulators.
These manipulator concepts can cover a wide range of motions while benefiting from the self-lock structure's properties, such as compactability, light weight, conserving energy, high-speed rotational motion, and large moment.

Each manipulator is discussed through the results of a kinematic simulation in which the angles of the involved origami joints are swept through a range of angles in order to obtain a desired motion.
Dynamics, including gravity, are outside the scope of this section, as are the details of any particular actuation scheme.
The rotational and translational manipulators were simulated for different values of the reduced central angle $\alpha$, and the trade-offs involved in choosing a particular value are discussed.

\subsection{Rotational Manipulator}

\begin{figure}
\centering
\includegraphics[width=\linewidth]{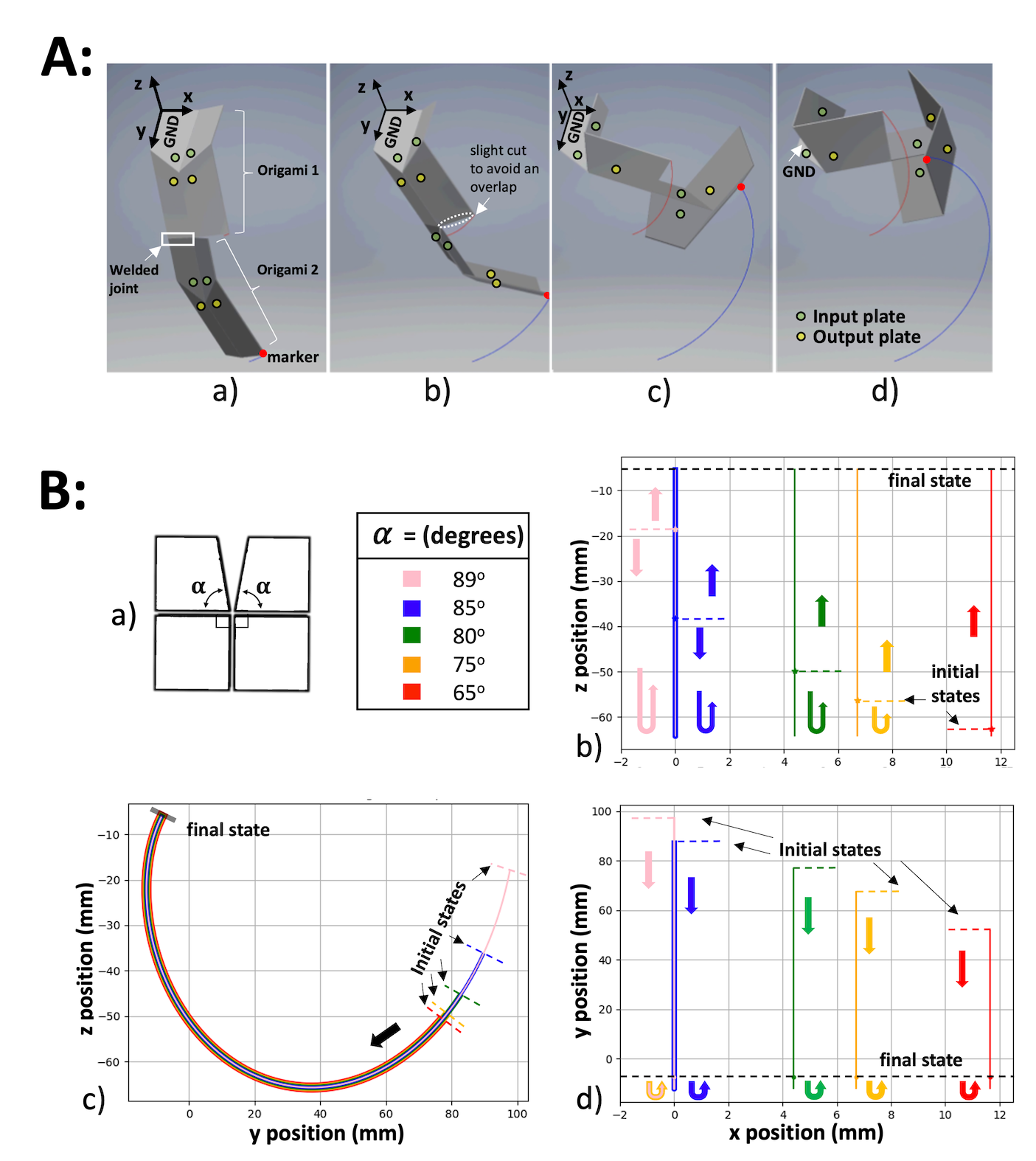}
\caption{
{\bf A}: Folding simulation of a rotational manipulator comprising two origami joints with $\alpha = 80^\circ$. The input plates and output plates are marked with green and yellow circles respectively.\\
{\bf B}:
a) The origami plates that are used in the manipulator and table of $\alpha$ values used for the two Self-Lock Origami units making up the manipulator.\\
b-d) Motion comparison of manipulator's end-effector (marker) for different values of the central angle $\alpha$, starting in the origami's semi-flat state and ending in the maximum possible fold state (see text).
} \label{rotationalManipulator}
\end{figure}

The rotational manipulator shown in figure \ref{rotationalManipulator} consists of two origami units, connected by a welded joint between one of the output plates of the first origami and one of the input plates of the second origami.
Connecting them on more than one plate could jeopardise their mobility; the connection method described here leaves each origami unit with a single degree of freedom.
There is a possibility of a collision between the first origami's output plates and the second origami's input plates while the manipulators move.
To avoid this, the plates are cut at a slight angle in those problematic areas (figure \ref{rotationalManipulator}Ab).
Cutting them does not affect the manipulator's movements because the plates' central angles are unaffected.
The motion of the manipulator is given in terms of the position of a designated end effector on one corner of the second origami's output plate, indicated in figure \ref{rotationalManipulator}A with the red dot labelled ``marker''.

Both of the origami making up the manipulator are constructed in the downwards configuration.
Figure \ref{rotationalManipulator}A gives the coordinate system of manipulators and shows the curling motion in 3D.
Figure \ref{rotationalManipulator}B shows the planar projection of this movement onto the $x$-$z$, $y$-$z$, and $x$-$y$ planes.
This manipulator is designed to produce a simple rotation about the $x$ axis, so the end effector does not move in this direction.
However, due to the change in the geometry, the end effector moves further from the $x = 0mm$ line with decreasing $\alpha$.
The greater the reduction of the central angles (i.e. the further $\alpha$ is from $90^\circ$), the shorter the range of motion on the y-axis.
The $89^\circ$ and $85^\circ$ degrees manipulators have similar and very close rotation around $x = 0mm$ due to their very small central angle cuts.
Due to defining the common MPF state at output angle $\gamma \coloneqq 36.5^\circ$ for all manipulator models, their final states and positions in all plots are the same.

The five manipulators with different values of $\alpha$ follow the overall trajectory but with a different initial state, so they all have the same range of motion on the $z$-axis.
The larger their central angle cuts are, the bigger the inaccessible configuration area near their flat state.
Therefore, their semi-flat configuration (initial state) starts farther away in the range of motion.
The movements of different rotational manipulator models in the 3D space are demonstrated in figure S8 in the supplementary material.

\subsection{Translational Manipulator}

\begin{figure}
\centering
\includegraphics[width=\linewidth]{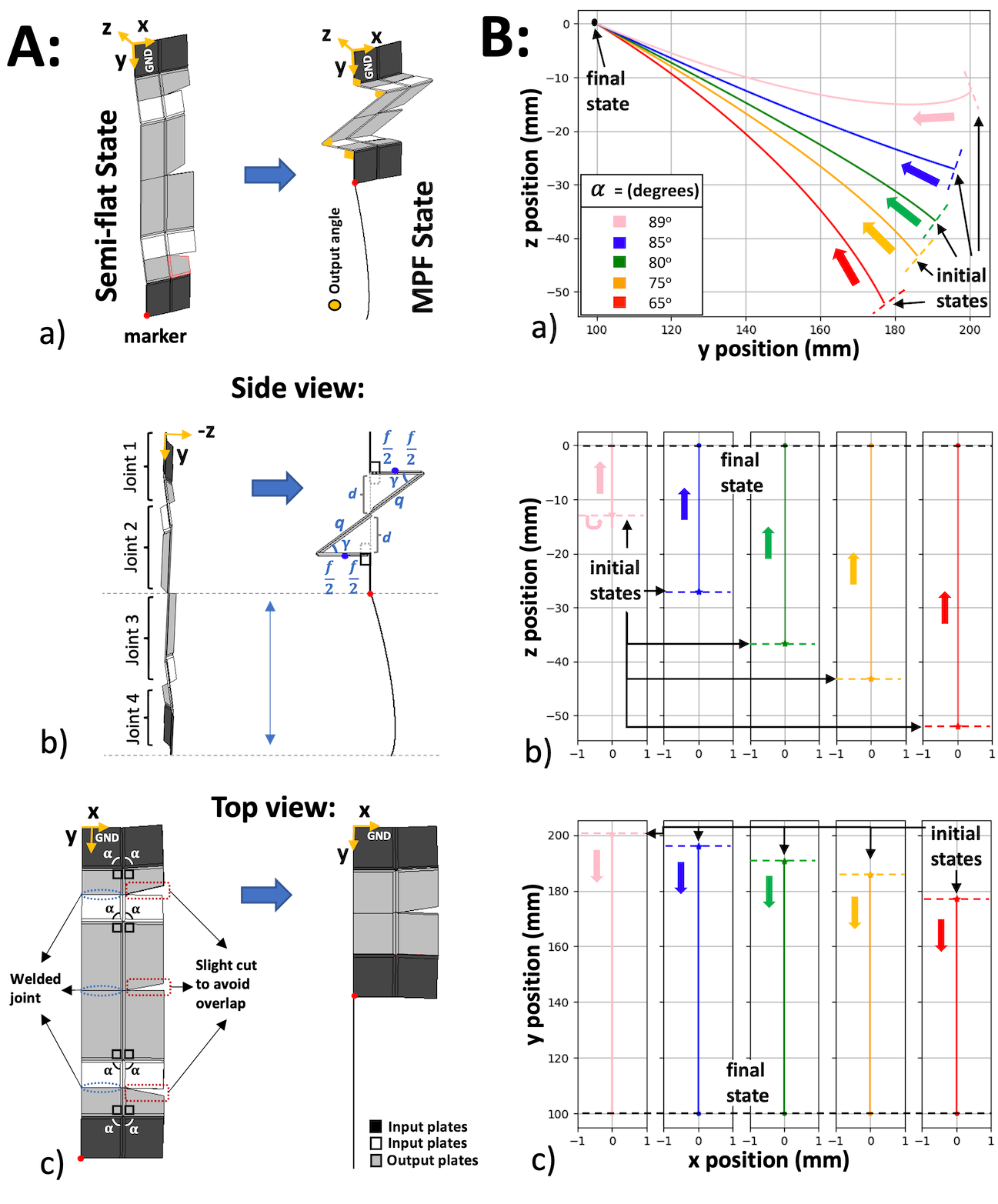}
\caption{
{\bf A}: The $89^\circ$ translational manipulator, comprising four separate Self-Lock Origami units. Two units are in the up configuration (black), and two are in the down configuration (white). Gray plates have central angles of $\frac\pi2$ and are shared between two origami in opposite configurations.
The output angles of the four origami units are marked with gold dots on the right figure.\\
{\bf B}:
a) Top view of the same motion as A. Weld joints and cuts made to avoid overlap between plates are labelled.
b) Side view of $89^\circ$ degrees manipulator's initial and final state, with the four origami joints indicated individually. Also indicated are the dimensions of the symmetrical zigzag pattern $f$, $q$, $d$, and $\gamma$.
{\bf C}: Two-dimensional workspace of translational manipulators with different values of $\alpha$. Initial and final states are labelled, and direction of motion is indicated with an arrow. They have separate initial states and common final state positions.
} \label{TranslationalManipulator}
\end{figure}

The translational manipulator shown in figure \ref{TranslationalManipulator} is constructed from four separate origami units alternating downward and upward configurations.
This is necessary in order to enable the ``zigzag'' state depicted in figure \ref{TranslationalManipulator}Aa.
To obtain the closest possible motion to a linear translational movement, the manipulator's geometry is required to be symmetrical.
Figure \ref{TranslationalManipulator}Ac shows the location of these connections and cuts.

Similar to the rotational manipulator, the origamis are connected on only one of their plates to retain the full number of degrees of freedom, and some perimeter cuts are made to avoid overlapping between the origamis' movements.
This manipulator is the only case where plate length deviates from the previously fixed 25mm, as indicated in figure \ref{TranslationalManipulator}Ab.
The first and last plates have length 25mm, but the lengths of other plates were derived geometrically.
The MPF angle $\gamma \coloneqq 36.5^\circ$ and the desired height $d = 25mm$ of the origami in the MPF state are used to find the plate lengths $f = d \cot \gamma$ and $q = d \sec \gamma$.
 
Figure \ref{TranslationalManipulator}A shows the initial and final states of the $89^\circ$ manipulator's translational movement in different views.
All origami units start in a semi-flat state and must be actuated simultaneously in order to achieve the depicted straight-line motion.
The first and the last joint are folded until reaching a $90^\circ$ output angle.
The second and third joints' final configurations are MPF.
Therefore, as for the rotational manipulator, varying $\alpha$ affects the initial but not the final position.

The two-dimensional plots of the manipulators' movements using a marker and defined reference frame are presented in figure \ref{TranslationalManipulator}C.
The main translational movement occurs along the $y$ axis, but the end effector does also move somewhat in the $z$ axis.
In the $y$-$z$ plane depicted in figure \ref{TranslationalManipulator}Ba, the $89^\circ$ manipulator stays closest to the $z = 0mm$ line.
The farther the origami models get from $\alpha = 90^\circ$, the further their semi-flat state gets from flat. As a result, the range of motion in the $z$ axis increases and the $y$ axis decreases, and their movements changes from a translational to a somewhat more rotational motion.
Among all the translational manipulators, the $89^\circ$ model produces the closest to ideal translational movement.

\subsection{Modular Manipulator}

Both manipulators could be combined or their constituent origami units could be actuated in other permutations in order to obtain a combination of translational and rotational motions which could be used for complex tasks.
Using the origami design in a modular structure could provide the opportunity to create and mimic traditional robots' complex movements.
This could be useful in exploration, where various motions are required in a limited space.
This section explores the possibility of a modular manipulator which combines Self-Lock Origami units in an arbitrary way.

\begin{figure}
\centering
\includegraphics[width=\linewidth]{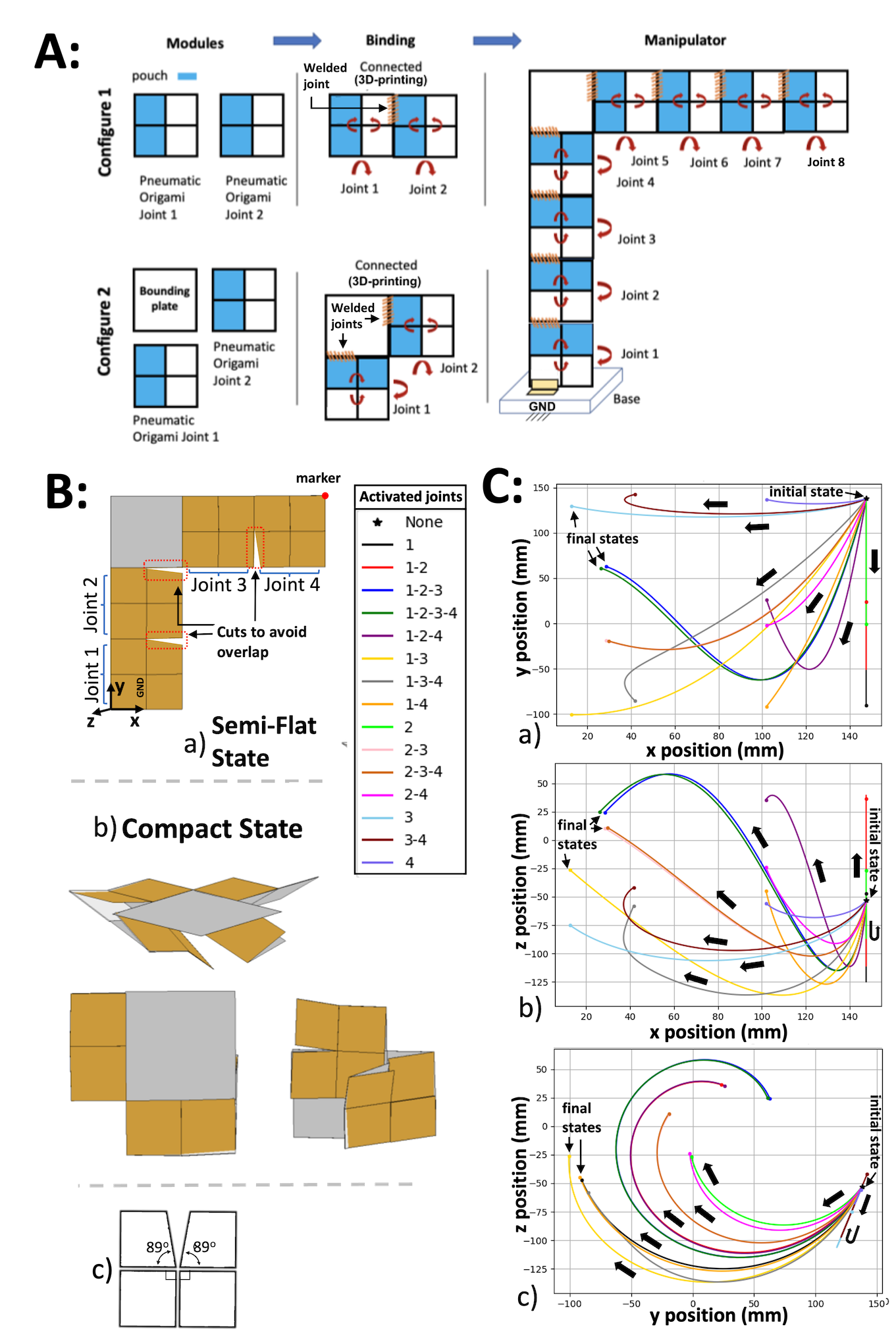}
\caption{
{\bf A:} Construction of the modular manipulator.
Designing modules phase: creating origami and the bounding plate (Configurations 1 and 2).
Binding phase: connecting the origami to each other and the bounding plate.
Manipulator: assembled manipulator with eight origami joints installed on a grounded base.
{\bf B:} a) Top view of the modular manipulator's semi-flat state with a square bounding plate and two origami connected to two sides of it.
b) Compact state: Top, bottom, and side views of the maximum possible fold (MPF) state.
c) $89^\circ$ origami plates are used in the manipulator.
{\bf C:} Two-dimensional plots of manipulator's movements starting in a semi-flat state as different combinations of joints are activated (see legend). Initial and final states are labeled, and direction of motion is indicated with an arrow.
} \label{modularMani}
\end{figure}

A modular manipulator can be built up from Self-Lock Origami units using two different basic construction cells, depicted in figure \ref{modularMani}A.
The first consists of two origami connected directly to each other with welded joints and small cutouts to avoid self-collision as discussed previously.
The second connects two origami via a rigid ``bounding plate'' interposed between them. 
This enables connecting the origami plates at different angles to change or increase the manipulator's workspace.
Bounding plates could be designed in different shapes to help the origami robots to explore different locations, but in the present work only considers a square plate equal in size to the origami units.
In the binding phase, constraints are defined for the movements in 3 directions or welded joints to create configurations 1 and 2.
Then, the last origami joint is connected to a base plate which serves as the ground link.

An example of such a modular manipulator made of $89^\circ$ degrees origami joints is presented in figure \ref{modularMani}.
By defining different rotational motions on the origami joints, the manipulator could take various shapes in the simulation.
Figure \ref{modularMani}B shows the manipulator's compact and semi-flat states with two $89^\circ$ origami connected at a right angle using a square bounding plate.
In the semi-flat state, none of the origami have been activated, and the manipulator is in the closest possible configuration to the flat state.
In the compact state, all joints have been activated and folded to the MPF configuration to minimise the overall size of the manipulator.
Figure \ref{modularMani}Ba additionally indicates the location of the end effector and the coordinate system.

The plots in figure \ref{modularMani}C present various possible trajectories which could be taken by this modular manipulator in each of the two-dimensional projections.
Each trajectory corresponds to a different sequence of joint activation, indicated in the legend as a sequence of joint numbers.
Each joint in the sequence folds from the semi-flat to MPF state, with each fold stopping early if a collision occurs.
For instance, activated joints 1-2 refers to folding of joint 1 and then folding joint 2 without overlapping joint 1.
The joint activation orders 1-2-3 and 1-2-3-4 have similar motion since after joints 1, 2, and 3 are activated, joint 4 does not have much space for folding.

Across the possible joint folding orders shown, the marker exhibits a large range of motion and a variety of different directions of motion in all three planes.
If only the joints on one side of the bounding plate are activated, the end effector trajectory is a a straight line.
The joint combination 1-2 generates straight-line movement in both the $x$-$y$ and $x$-$z$ planes, whereas the joint combination 3-4 produced straight-line motion in the $y$-$z$ plane.

One of the advantages of using origami in modular manipulators is the ability to convert them into structures with small volumes due to the joints' geometries.
Manipulators can achieve this by moving their origami into MPF or semi-flat states in different directions.
Also, depending on the application, some of the manipulator's joints can be excluded in order to produce different manipulators.
This can even be enforced temporarily, without modifying the manipulator, by simply folding the joints using their actuators and holding them in that state.


\section{Conclusions}\label{conclusions}

Origami structures could be a great substitute for traditional joints where space limitation is the main issue. They could adapt their size and shape based on the available space.
These computer simulations and mathematical models illustrate the origami structure's high performance. In different origami states, their motion's speed and output moment are more significant compared to a simple fold with the same actuator.
Increasing the thickness of the origami plates would increase the mechanical strength of the structure, but decrease its flexibility. On the other hand, increasing the soft joint length between the plates increases flexibility while decreasing strength. The trade-off between these characteristics is a potential direction of future research.

As proof of concept, different types of manipulators have been developed and simulated using the origami joints. They could produce a variety of rotational and translational motions depending on their origami's central angles and configurations.
The proposed joint could potentially have many industrial applications, especially in space where the weight and volume of the tools matter the most. It could generate various rotational and translational motions depending on the central angle while maintaining a very lightweight and low fabrication cost due to the flat-foldability properties (simplicity of sheet-like structures for mass production).

One limitation of this work is the assumption of a revolute joint in the design's analytical modelling. This could result in discrepancies between the experimental and the modelling results. Future work will focus on the development of methods and designs for soft joint production to mitigate the unpredictability of soft materials.
This paper has made the assumption that joint angles can be set to a desired value; for physical prototypes, a control scheme is required which combines the kinematic model with sensing and feedback based on the desired end-effector position of the manipulator.
Finally, future work will develop methods and experiments to take advantage of this origami joint for real-life applications.


\section*{Author Disclosure Statement}
No competing financial interests exist.
This work was not supported by any organisation.



\section{Nomenclatures}
\begin{tabular}{p{1.1cm}p{6.7cm}}
    $S$ & Angle subtended by one arc-shaped side of the inflated pouch actuator. \\
    $\theta_2$ & Angular deviation of plates 2 and 3 from coplanar. \\
    $\theta_3$ & Angular deviation of plates 3 and 4 from coplanar. \\
    $\alpha$ & Central angle of plate 1 of the origami, here set to $\alpha$. \\
    $\alpha_{12}$ & Central angle of plate 2 of the origami, here set to $\alpha$. \\
    $\alpha_{34}$ & Central angle of plate 3 of the origami, here always equal to $\frac{\pi}{2}$. \\
    $\alpha_{41}$ & Central angle of plate 4 of the origami, here always equal to $\frac{\pi}{2}$. \\
    $n$ & Distance from the corner of the plate where the corner of the pouch actuator meets the edge of the plate. \\
    $P$ & Fixed pressure within the pouch actuator. \\
    $L_p$ & Half-length of the pouch actuator when flat, equal. \\
    $d$ & Height of an origami cell in its MPF state in the translational manipulator. \\
    $\theta_1$ & Input angle, the angular deviation of plates 1 and 2 from coplanar. \\
    $M_\text{input}$ & Input moment created directly by the pouch motor. \\
    $L$ & Length of the inflated pouch actuator. \\
    $L_0$ & Length of the pouch actuator when flat, equal to $2L_p$. \\
    $L_1$ & Length removed from the far side of plates 1 and 2 by cutting. \\
    $m$ & Length of the sides of the input plate before cutting. \\
    MPF & Maximum Possible Fold, the state where $\theta_4 = \gamma$. \\
    MA & Mechanical advantage of the origami structure, defined as $M_\text{output} / M_\text{input}$. \\
    $\theta_4$ & Output angle, the angular deviation of plates 1 and 4 from coplanar. \\
    $M_\text{output}$ & Output moment created by the pouch motor through the mechanical advantage of the origami. \\
    $\alpha_{12}$ & Reduced value of central angles $\alpha_{12}$ and $\alpha_{23}$. \\
    $\gamma$ & Value of $\theta_4$ in the MPF state. \\
    $f$ & Width of the horizontal components of the "zigzag" configuration of the translational manipulator. \\
    $D$ & Width of the pouch actuator when flat. \\ 
\end{tabular}


\begin{thebibliography}{99}

\bibitem{jasim2018origami}
Jasim B and Taheri P.
\newblock An origami-based portable solar panel system.
\newblock In \emph{2018 IEEE 9th Annual Information Technology, Electronics and
  Mobile Communication Conference (IEMCON)}. IEEE, pp. 199--203.

\bibitem{tang2014origami}
Tang R, Huang H, Tu H et~al.
\newblock Origami-enabled deformable silicon solar cells.
\newblock \emph{Applied Physics Letters} 2014; 104(8): 083501.

\bibitem{arya2017crease}
Arya M, Lee N and Pellegrino S.
\newblock Crease-free biaxial packaging of thick membranes with slipping folds.
\newblock \emph{International Journal of Solids and Structures} 2017; 108:
  24--39.

\bibitem{zhai2020situ}
Zhai Z, Wang Y, Lin K et~al.
\newblock In situ stiffness manipulation using elegant curved origami.
\newblock \emph{Science advances} 2020; 6(47): eabe2000.

\bibitem{quaglia2014balancing}
Quaglia C, Yu N, Thrall A et~al.
\newblock Balancing energy efficiency and structural performance through
  multi-objective shape optimization: Case study of a rapidly deployable
  origami-inspired shelter.
\newblock \emph{Energy and Buildings} 2014; 82: 733--745.

\bibitem{ye2022novel}
Ye K and Ji J.
\newblock A novel morphing propeller system inspired by origami-based
  structure.
\newblock \emph{Journal of Mechanisms and Robotics} 2022; 15(1): 011006.

\bibitem{dai2010origami}
Dai J and Caldwell D.
\newblock Origami-based robotic paper-and-board packaging for food industry.
\newblock \emph{Trends in food science \& technology} 2010; 21(3): 153--157.

\bibitem{onal2011towards}
Onal CD, Wood RJ and Rus D.
\newblock Towards printable robotics: Origami-inspired planar fabrication of
  three-dimensional mechanisms.
\newblock In \emph{2011 IEEE international conference on robotics and
  automation}. IEEE, pp. 4608--4613.

\bibitem{zhakypov2015design}
Zhakypov Z, Falahi M, Shah M et~al.
\newblock The design and control of the multi-modal locomotion origami robot,
  tribot.
\newblock In \emph{2015 IEEE/RSJ International Conference on Intelligent Robots
  and Systems (IROS)}. IEEE, pp. 4349--4355.

\bibitem{yang2017smartphone}
Yang JS, Shin J, Choi S et~al.
\newblock Smartphone diagnostics unit (sdu) for the assessment of human stress
  and inflammation level assisted by biomarker ink, fountain pen, and origami
  holder for strip biosensor.
\newblock \emph{Sensors and Actuators B: Chemical} 2017; 241: 80--84.

\bibitem{kimionis20153d}
Kimionis J, Isakov M, Koh BS et~al.
\newblock 3d-printed origami packaging with inkjet-printed antennas for rf
  harvesting sensors.
\newblock \emph{IEEE Transactions on Microwave Theory and Techniques} 2015;
  63(12): 4521--4532.

\bibitem{norton2008design}
Norton RL.
\newblock \emph{Design of machinery: an introduction to the synthesis and
  analysis of mechanisms and machines}.
\newblock McGraw-Hill/Higher Education, 2008.

\bibitem{rus2018design}
Rus D and Tolley MT.
\newblock Design, fabrication and control of origami robots.
\newblock \emph{Nature Reviews Materials} 2018; 3(6): 101--112.

\bibitem{mehta2014cogeneration}
Mehta AM, DelPreto J, Shaya B et~al.
\newblock Cogeneration of mechanical, electrical, and software designs for
  printable robots from structural specifications.
\newblock In \emph{2014 IEEE/RSJ International Conference on Intelligent Robots
  and Systems}. IEEE, pp. 2892--2897.

\bibitem{mehta2014end}
Mehta AM and Rus D.
\newblock An end-to-end system for designing mechanical structures for
  print-and-fold robots.
\newblock In \emph{2014 IEEE International Conference on Robotics and
  Automation (ICRA)}. IEEE, pp. 1460--1465.

\bibitem{kohut2013precise}
Kohut NJ, Pullin AO, Haldane DW et~al.
\newblock Precise dynamic turning of a 10 cm legged robot on a low friction
  surface using a tail.
\newblock In \emph{2013 IEEE International Conference on Robotics and
  Automation}. IEEE, pp. 3299--3306.

\bibitem{haldane2013animal}
Haldane DW, Peterson KC, Bermudez FLG et~al.
\newblock Animal-inspired design and aerodynamic stabilization of a hexapedal
  millirobot.
\newblock In \emph{2013 IEEE International Conference on Robotics and
  Automation}. IEEE, pp. 3279--3286.

\bibitem{zhakypov2018design}
Zhakypov Z and Paik J.
\newblock Design methodology for constructing multimaterial origami robots and
  machines.
\newblock \emph{IEEE Transactions on Robotics} 2018; 34(1): 151--165.

\bibitem{rossiter2014kirigami}
Rossiter J and Sareh S.
\newblock Kirigami design and fabrication for biomimetic robotics.
\newblock In \emph{Bioinspiration, Biomimetics, and Bioreplication 2014},
  volume 9055. SPIE, pp. 105--112.

\bibitem{firouzeh2017grasp}
Firouzeh A and Paik J.
\newblock Grasp mode and compliance control of an underactuated origami gripper
  using adjustable stiffness joints.
\newblock \emph{Ieee/asme Transactions on Mechatronics} 2017; 22(5):
  2165--2173.

\bibitem{geckeler2022bistable}
Geckeler C and Mintchev S.
\newblock Bistable helical origami gripper for sensor placement on branches.
\newblock \emph{Advanced Intelligent Systems} 2022; 4(10): 2200087.

\bibitem{chan2017design}
Chan YH, Tse Z and Ren H.
\newblock Design evolution and pilot study for a kirigami-inspired flexible and
  soft anthropomorphic robotic hand.
\newblock In \emph{2017 18th international conference on advanced robotics
  (ICAR)}. IEEE, pp. 432--437.

\bibitem{suzuki2020origami}
Suzuki H and Wood RJ.
\newblock Origami-inspired miniature manipulator for teleoperated microsurgery.
\newblock \emph{Nature Machine Intelligence} 2020; 2(8): 437--446.

\bibitem{firouzeh2017under}
Firouzeh A and Paik J.
\newblock An under-actuated origami gripper with adjustable stiffness joints
  for multiple grasp modes.
\newblock \emph{Smart Materials and Structures} 2017; 26(5): 055035.

\bibitem{koh2012omega}
Koh JS and Cho KJ.
\newblock Omega-shaped inchworm-inspired crawling robot with
  large-index-and-pitch (lip) sma spring actuators.
\newblock \emph{IEEE/ASME Transactions On Mechatronics} 2012; 18(2): 419--429.

\bibitem{boyvat2017addressable}
Boyvat M, Koh JS and Wood RJ.
\newblock Addressable wireless actuation for multijoint folding robots and
  devices.
\newblock \emph{Science Robotics} 2017; 2(8): eaan1544.

\bibitem{salerno2016novel}
Salerno M, Zhang K, Menciassi A et~al.
\newblock A novel 4-dof origami grasper with an sma-actuation system for
  minimally invasive surgery.
\newblock \emph{IEEE Transactions on Robotics} 2016; 32(3): 484--498.

\bibitem{yi2018customizable}
Yi J, Chen X, Song C et~al.
\newblock Customizable three-dimensional-printed origami soft robotic joint
  with effective behavior shaping for safe interactions.
\newblock \emph{IEEE Transactions on Robotics} 2018; 35(1): 114--123.

\bibitem{sung2015foldable}
Sung C and Rus D.
\newblock Foldable joints for foldable robots.
\newblock \emph{Journal of Mechanisms and Robotics} 2015; 7(2): 021012.

\bibitem{taylor2019mr}
Taylor AJ, Slutzky T, Feuerman L et~al.
\newblock Mr-conditional sma-based origami joint.
\newblock \emph{IEEE/ASME Transactions on Mechatronics} 2019; 24(2): 883--888.

\bibitem{baek2020ladybird}
Baek SM, Yim S, Chae SH et~al.
\newblock Ladybird beetle--inspired compliant origami.
\newblock \emph{Science Robotics} 2020; 5(41): eaaz6262.

\bibitem{saito2017investigation}
Saito K, Nomura S, Yamamoto S et~al.
\newblock Investigation of hindwing folding in ladybird beetles by artificial
  elytron transplantation and microcomputed tomography.
\newblock \emph{Proceedings of the National Academy of Sciences} 2017; 114(22):
  5624--5628.

\bibitem{qiu2021design}
Qiu L, Yu Y and Liu Y.
\newblock Design and analysis of lamina emergent joint (lej) based on origami
  technology and mortise-tenon structure.
\newblock \emph{Mechanism and Machine Theory} 2021; 160: 104298.

\bibitem{zare2021design}
Zare S and Teodorescu M.
\newblock Design and analysis of plate angles of the four-vertex origami
  pattern and its impacts on movement of rotational joints.
\newblock \emph{Smart Materials and Structures} 2021; 30(9): 095012.

\bibitem{niiyama2015pouch}
Niiyama R, Sun X, Sung C et~al.
\newblock Pouch motors: Printable soft actuators integrated with computational
  design.
\newblock \emph{Soft Robotics} 2015; 2(2): 59--70.

\bibitem{brown2022approaches}
Brown NC, Ynchausti C, Lytle A et~al.
\newblock Approaches for minimizing joints in single-degree-of-freedom
  origami-based mechanisms.
\newblock \emph{Journal of Mechanical Design} 2022; 144(10): 103301.

\bibitem{faber2018bioinspired}
Faber JA, Arrieta AF and Studart AR.
\newblock Bioinspired spring origami.
\newblock \emph{Science} 2018; 359(6382): 1386--1391.

\bibitem{hull2002modelling}
Hull TC et~al.
\newblock Modelling the folding of paper into three dimensions using affine
  transformations.
\newblock \emph{Linear Algebra and its applications} 2002; 348(1-3): 273--282.

\bibitem{hull2002combinatorics}
Hull T.
\newblock The combinatorics of flat folds: a survey.
\newblock In \emph{Origami3: Proceedings of the 3rd International Meeting of
  Origami Science, Math, and Education}. pp. 29--38.

\bibitem{song2016microscale}
Song Z, Lv C, Liang M et~al.
\newblock Microscale silicon origami.
\newblock \emph{Small} 2016; 12(39): 5401--5406.

\bibitem{bowen2014position}
Bowen LA, Baxter W, Magleby SP et~al.
\newblock A position analysis of coupled spherical mechanisms found in action
  origami.
\newblock \emph{Mechanism and Machine Theory} 2014; 77: 13--24.

\bibitem{chiang1988kinematics}
Chiang CH.
\newblock \emph{Kinematics of spherical mechanisms}.
\newblock Cambridge University Press, 1988.

\bibitem{yang1964application}
Yang AT and Freudenstein F.
\newblock Application of dual-number quaternion algebra to the analysis of
  spatial mechanisms.
\newblock \emph{Journal of Applied Mechanics} 1964; 31(2): 300--308.

\bibitem{sun2015self}
Sun X, Felton SM, Niiyama R et~al.
\newblock Self-folding and self-actuating robots: A pneumatic approach.
\newblock In \emph{2015 IEEE International Conference on Robotics and
  Automation (ICRA)}. IEEE, pp. 3160--3165.

\bibitem{yang1965static}
Yang AT.
\newblock Static force and torque analysis of spherical four-bar mechanisms.
\newblock \emph{Journal of Engineering for Industry} 1965; 87(2): 221--227.

\end{thebibliography}
\end{document}